%% file: main.tex
\documentclass[conference,9pt]{IEEEtran}
\IEEEoverridecommandlockouts

\usepackage{cite}
\usepackage{amsmath,amssymb,amsfonts}
\usepackage{algorithmic}
\usepackage{graphicx}
\usepackage{textcomp}
\usepackage{xcolor}
\usepackage{verbatim}
\usepackage{framed}
\usepackage{listings}
\usepackage{multirow}
\usepackage{booktabs}
\usepackage{float}
\def\BibTeX{{\rm B\kern-.05em{\sc i\kern-.025em b}\kern-.08em
    T\kern-.1667em\lower.7ex\hbox{E}\kern-.125emX}}

\newcommand{\tabincell}[2]{\begin{tabular}{@{}#1@{}}#2\end{tabular}}

\begin{document}

\title{
Hybrid-NL2SVA: Integrating RAG and Finetuning for LLM-based NL2SVA
}

\author{%
{\normalsize Weihua Xiao, Derek Ekberg, Siddharth Garg, Ramesh Karri}\\
{\normalsize NYU Tandon School of Engineering, Brooklyn, NY, USA}\\
{\normalsize Emails: \{wx2356, dhe9940\}@nyu.edu, siddharth.j.garg@gmail.com, rkarri@nyu.edu}
}


\maketitle

\begin{abstract}

\textit{SystemVerilog Assertion}s (\textit{SVA}s) are critical for verifying the correctness of hardware designs, but manually writing them from natural language property descriptions, i.e., \textit{NL2SVA}, remains a labor-intensive and error-prone task. 
Recent advances in \textit{large language model}s (\textit{LLM}s) offer opportunities to automate this translation. 
However, existing models still struggle with understanding domain-specific syntax and semantics. 
To enhance LLM performance in NL2SVA, we propose a customized \textit{retrieval-augmented generation} (\textit{RAG}) framework and a synthetic fine-tuning dataset that together improve LLM's performance. 
Our RAG framework (i) constructs a context-preserving database via \textit{dynamic splitting technique}, (ii) combines global semantic retrieval with keyword-guided retrieval to extract SVA operator-related contexts via \textit{HybridRetrieval}, and (iii) validate and correct the use of SVA operators in LLM-generated SVAs via \textit{SVA operator-based rechecking}.
To further improve lightweight models over NL2SVA, our fine-tuning dataset provides \textit{prompt-guided explanation}s that teach LLMs the layer-by-layer construction process of concurrent SVAs, enabling supervised fine-tuning that greatly improves syntax and functionality accuracy.
To evaluate the performance of LLMs over NL2SVA, we construct the largest evaluation dataset for NL2SVA, comprising $40$ Verilog designs and $229$ formally verified SVAs with detailed annotations.
Experimental results show that our customized RAG framework increases the number of functionality matched SVAs by $58.42 \%$ over \textit{GPT-4o-mini}, while \textit{Qwen2.5-Coder-7B-Instruct} fine-tuned on our fine-tuning dataset and integrated with HybridRetrieval achieves a $59.05 \%$ over the base Qwen model.
\end{abstract}


\input{Tex/Introduction}
\input{Tex/Preliminary}

\input{Tex/Methodology}
\input{Tex/Experiments}

\input{Tex/Conclusion}
\bibliographystyle{IEEEtran}
\bibliography{Reference}
\onecolumn
\newpage
\appendix

\input{Tex/Appendix}

\end{document}

%% file: Tex/Introduction.tex
\section{Introduction}
\label{sec:introduction}
\textit{SystemVerilog Assertion}s (\textit{SVA}s) are essential tools in hardware verification, formally specifying expected design behaviors, namely the \textit{design properties}, and serving as embedded checkers that continuously validate the implementation against its specification~\cite{Witharana22,SystemVerilog18}.
However, writing SVAs manually is difficult, which consists of two sub-tasks~\cite{Vasudevan10,Witharana23}.
The first task is to extract intended properties, described in natural language, from hardware designs and detailed specification documents. 
Once these are identified, the second task is to implement these natural language properties as SVAs, i.e., the NL2SVA task. 
Recent advances in \textit{Natural Language Processing} (\textit{NLP}) or \textit{Large Language Models} (\textit{LLMs}), such as \textit{GPT}-$4$\textit{o} and \textit{DeepSeek}, have shown promising potential in automating both sub-tasks~\cite{Radu24,Mali24}.

For the first sub-task of property extraction, several recent works have explored different techniques~\cite{Yan25,parthasarathy2021spectosva,Meng24}.
\textit{AssertLLM}~\cite{Yan25} proposes a multi-LLM framework that processes the complete specification document to extract design properties for each architectural signal.
\textit{SpecToSVA}~\cite{parthasarathy2021spectosva} trains a machine-learning-based sentence classifier to automatically detect property-relevant sentences from specification documents, reducing manual effort in identifying verification properties.
\textit{NSPG}~\cite{Meng24} introduces a fine-tuned domain-specific LLM to  mine hardware security properties from specification documents.


Recent works have also explored automating the NL2SVA sub-task using both \textit{rule-based} and \textit{LLM-based} methods. Traditional rule-based methods, such as \textit{GLaST}~\cite{Harris16} and \textit{Ease}~\cite{Krishnamurthy19}, rely on manually crafted grammars, rules, or templates, but struggle to generalize across diverse specifications and require substantial manual effort. 
More recently, LLM-based approaches like \textit{nl2spec}~\cite{Cosler23} introduce interactive prompting frameworks where engineers refine LLM-suggested subformulas to improve accuracy, but it depends heavily on human intervention and careful prompt design. 
AssertLLM also proposes a multi-LLM pipeline and incorporates a \textit{retrieval-augmented generation} (\textit{RAG}) component to assist NL2SVA; 
however, it just uses the generic retrieval method, and focuses only on syntax correctness and \textit{Formal Property Verification} (\textit{FPV}) pass of generated SVAs, without evaluating functionality match of generated SVAs to the natural language property description.
In~\cite{wu2025}, the authors first convert the natural language description into an intermediate formal representation, and then decompose that representation into fragments, translates each fragment into its corresponding SVA, and finally combines them into a complete assertion.
\cite{Shahidzadeh24} uses pairs of SVAs and natural language explanations to finetune \textit{GPT-3.5-Turbo}.
While demonstrating promise, these methods have struggled to achieve high accuracy, often failing to translate a desired natural language specification into the SVA. We argue that this is primarily because existing methods have not properly customized their RAG and/or finetuning pipelines to the task at hand.

In this paper, we propose a \textit{customized RAG framework} and a \textit{synthetic fine-tuning dataset} to improve LLM-based NL2SVA. 
The main contributions are as follows:
\begin{itemize}
    \item [(1)] We propose the first systematic framework to customize traditional RAG for NL2SVA.
    \item [(2)] We introduce a \textit{dynamic splitting technique} for database construction to preserve semantic context.
    \item [(3)] We develop \textit{HybridRetrieval} to improve retrieval relevance by combining global semantic and keyword-guided retrievals.
    \item [(4)] We propose an \textit{SVA operator-based rechecking} to refine and correct LLM-generated assertions.
    \item [(5)] We provide a synthetic fine-tuning dataset of \textit{prompt-guided explanation}s for the layer-by-layer construction of SVAs.
    \item [(6)] We construct the largest evaluation dataset for NL2SVA.
\end{itemize}

In the remainder of this paper, Section~\ref{sec:Preliminary} introduces some preliminaries about SVAs and RAG. 
Section~\ref{subsec:RAGSVAG} presents the proposed customized RAG framework, including three key components: the dynamic splitting technique for database construction, the HybridRetrieval method for improving retrieval process, and the SVA operator-based rechecking flow for SVA refinement. 
Section~\ref{subsec:synthetic-finetuning-dataset} introduces our proposed synthetic fine-tuning dataset.
Section~\ref{subsec:Evaluation-Dataset} introduces our evaluation dataset. 
Then, the experimental results are reported in Section~\ref{sec:exp}. 
Finally, Section~\ref{sec:con} concludes the paper.

%% file: Tex/Preliminary.tex
\section{Preliminary and Related Works}
\label{sec:Preliminary}
\subsection{SystemVerilog Assertions}
\label{subsec:SVA}
An SVA is constructed using different signals from the hardware design, that are then composed using various \textit{SVA operator}s to describe functional or temporal relationships between signals.
In SystemVerilog, there are two kinds of assertions: \textit{immediate assertions} and \textit{concurrent assertions}~\cite{IEEEStd}.

Immediate assertions execute  within procedural code (for example, inside an \verb|initial|/\verb|always| block). 
They check combinational conditions using \textit{Boolean operator}s such as \texttt{!}, \texttt{\&\&}, and \texttt{||}, serving as runtime self-checks during simulation.
For example:
\begin{verbatim}
     assert (valid && (req || grant));
\end{verbatim}
It checks the combinational condition that signal \verb|valid| is true, and at least one of signal \verb|req| or signal \verb|grant| is true.

\begin{figure}
    \centering
    \includegraphics[width=0.65\linewidth]{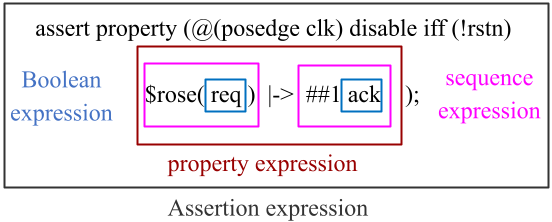}
    \caption{Four layers of an example concurrent assertion.}
    \label{fig:four-layers-sva}
\end{figure}

A concurrent assertion used to specify temporal properties of a design is constructed in four hierarchical layers~\cite{IEEEStd}. 
\begin{itemize}
    \item \textit{Boolean Layer}: that combines signals using Boolean operators, as noted above.
    \item \textit{Sequence Layer}: that applies \textit{Sequence Operator}s to one or more combinational expressions obtained from the Boolean layer. Sequence operators are shown in Table~\ref{tab:sequence_sva_operators}.
    \item \textit{Property Layer}: that applies \textit{Property Operator}s, shown in Table~\ref{tab:property_sva_operators}, to one or more sequential expressions.
    \item \textit{Verification Layer}: that binds a property expression to the design’s clock and reset signals, and wrap it in an \texttt{assert property (...)} directive.
\end{itemize}
Each SVA operator may take one or two operands. An operator introduced at a layer accepts operands that are expressions of that layer or of the layer immediately below it. Table~\ref{tab:sequence_sva_operators} shows $7$ types of sequence SVA operators and Table~\ref{tab:property_sva_operators}  shows $3$ types of property SVA operators.
Fig.~\ref{fig:four-layers-sva} shows an example of an SVA, where all expressions in the four layers are marked using boxes with different colors.
\begin{table}[t]
\centering
\caption{Sequence SVA Operators}
\label{tab:sequence_sva_operators}
\begin{tabular}{|c|c|}
\hline
\textbf{SVA Operator} & \textbf{Explanation} \\ \hline
\texttt{\#\#N s} & \tabincell{l}{The evaluation of sequence expression \texttt{s} is delayed\\ by \texttt{N} clock cycles.} \\ \hline
\texttt{\$rose(s)} & \tabincell{l}{Returns $1$ if the LSB of sequence expression \texttt{s}\\ changed to $1$. Otherwise, returns $0$.} \\ \hline
\texttt{\$fell(s)} & \tabincell{l}{Returns $1$ if the LSB of sequence expression \texttt{s}\\ changed to $0$. Otherwise, returns $0$.} \\ \hline
\texttt{\$past(s,N)} & \tabincell{l}{Returns the value of sequence \texttt{s} in a \texttt{N} clock cycle \\step prior to the current one.} \\ \hline
\texttt{\$stable(s)} & \tabincell{l}{Returns $1$ if the value of sequence expression \texttt{s}\\ did not change. Otherwise, returns $0$.} \\ \hline
\texttt{\$onehot(s)} & \tabincell{l}{Returns $1$ if one bit of sequence expression \texttt{s}\\ is $1$. Otherwise, returns $0$.} \\ \hline
\texttt{\$onehot0(s)} & \tabincell{l}{Returns $1$ if no more than one bit of sequence\\ expression \texttt{s} is $1$. Otherwise, returns $0$.} \\ \hline
\end{tabular}
\end{table}

\begin{table}[t]
\centering
\caption{Property SVA Operators}
\label{tab:property_sva_operators}
\begin{tabular}{|c|c|}
\hline
\textbf{SVA Operator} & \textbf{Explanation} \\
\hline
\texttt{s |-> p} & \tabincell{l}{for every match of the sequence expression \texttt{s} \\beginning at the start point, the evaluation of\\ property expression \texttt{p} beginning in the current\\ clock cycle at the end point of the match \\succeeds and returns $1$.} \\ \hline
\texttt{s |=> p} & \tabincell{l}{for every match of the sequence expression \texttt{s} \\beginning at the start point, the evaluation of\\ property expression \texttt{p} beginning in the next\\ clock cycle at the end point of the match \\succeeds and returns $1$.} \\ \hline
\texttt{s\_eventually p} & \tabincell{l}{Return $1$ if there exists a current or future clock \\cycle at which property expression \texttt{p} is $1$.} \\ \hline
\end{tabular}
\end{table}

\subsection{Retrieval-Augmented Generation}
\label{subsec:RAG}
RAG is a technique that enhances the performance of LLMs by incorporating relevant external knowledge~\cite{Lewis20}.
Instead of relying solely on the LLM's internal knowledge, RAG augments the model’s input by retrieving relevant external documents from a knowledge database based on the query.
RAG has been successfully applied in different fields, such as code generation~\cite{Koziolek24}, medical QA~\cite{xiong2024}, financial QA~\cite{Setty24}, and conversational agents~\cite{alonso2024}.

RAG methods leverage a specialized database of \textit{vector embeddings}, created by first splitting the source materials, such as documents or code repositories, into smaller \textit{chunks} and then transforming each chunk into a numerical representation using an embedding model, i.e., a vector embedding. 
The strategy used for splitting the source materials significantly impacts the relevance of retrieved chunks to the input prompt. 
A common practice is to apply \textit{static splitting}, where the source data is divided into fixed-size text chunks, typically based on a constant number of characters. 
A common method for enhancing the static splitting techniques is to employ LLM-based splitting, which leverages LLMs to identify more semantically coherent chunk boundaries~\cite{singh24,chang24n,dong24}.
However, they require careful prompt design and substantial computational overhead.

The \textit{retrieval process}, which serves as the second key component of RAG, begins with an input prompt.
This prompt is transformed into a vector embedding using the same embedding model for constructing the database.
Then, the retrieval process uses \textit{similarity search} to identify the top-ranked or most similar chunks that align with the the input prompt, returning them as the \textit{relevant context}.
The original input prompt is then augmented with the relevant context to form a \textit{combined prompt}, enriching the LLM with domain-specific knowledge. 
The combined prompt guides the LLM to generate more accurate and context-aware responses. 




\subsection{Related Works}
A critical component for evaluating LLM-based
NL2SVA is the evaluation dataset. 
Prior works such as \textit{AssertionBench}~\cite{Vaishnavi25}, \textit{AssertLLM}~\cite{Yan25}, and \textit{FVEval}~\cite{Minwoo24} have provided open-source
datasets for benchmarking assertion generation. 
However, AssertionBench and AssertLLM lack detailed contextual information, such as natural language explanations of the SVAs.
This missing information is essential for rigorously evaluating
the functionality match between the generated SVAs and their
expected natural language property descriptions. 
FVEval introduces a more comprehensive dataset, containing $13$ hardware
designs, $80$ SVAs with corresponding human-written property
explanations.

%% file: Tex/Methodology.tex
\section{Methodology}
This section will introduce the main work of this paper.
In Section~\ref{subsec:RAGSVAG}, it introduces the proposed customized RAG framework for NL2SVA.
In Section~\ref{subsec:synthetic-finetuning-dataset}, it introduces the proposed fine-tuning dataset with prompt-guided explanations.
In Section~\ref{subsec:Evaluation-Dataset}, it introduces the proposed evaluation dataset.

\subsection{Customized RAG Framework for NL2SVA}
\label{subsec:RAGSVAG}
\begin{figure}
    \centering
    \includegraphics[width=\linewidth]{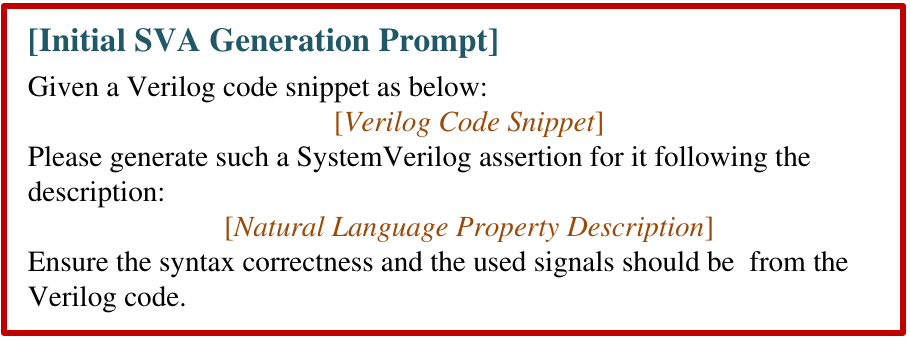}
    \caption{Initial SVA generation prompt.}
    \label{fig:initial_sva_generation_prompt}
\end{figure}

\begin{figure}
    \centering
    \includegraphics[width=\linewidth]{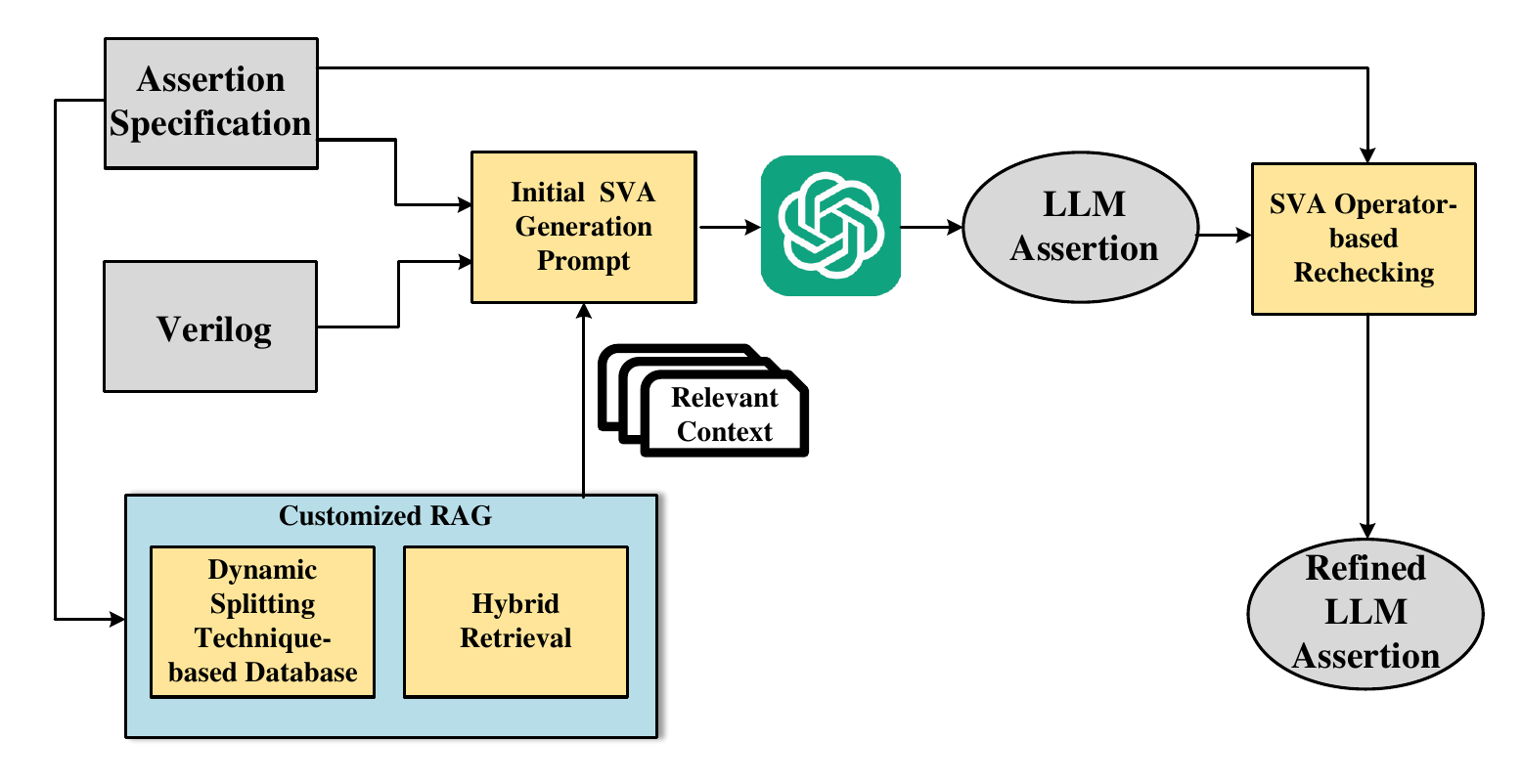}
    \caption{
    Overall flow of the proposed customized RAG framework: apply the proposed dynamic splitting to construct the database; apply the proposed HybridRetrieval to extract relevant context; apply the proposed SVA operator-based rechecking for correcting errors in generated SVAs.
    }
    \label{fig:overallflow-RAGSVAG}
\end{figure}

The overall flow is illustrated in Fig.~\ref{fig:overallflow-RAGSVAG}. 
Given a Verilog design and a natural language assertion specification, the framework first applies a customized RAG to retrieve relevant contextual information based on the input specification.
The customized RAG component is built upon two key techniques: 
\textit{dynamic splitting} for constructing a code-based retrieval database from SystemVerilog textbooks (Section~\ref{subsubsec:Dynamic-Splitting}), and \textit{HybridRetrieval} that enhances the retrieval  by integrating global semantic retrieval with keyword-guided retrieval (Section~\ref{subsubsec:HybridRetrieval}). 
The retrieved information is incorporated into the \textit{Initial Input Prompt} to assist in generating a better initial SVA. 
The initial input prompt is shown in Fig.~\ref{fig:initial_sva_generation_prompt}.



To improve correctness of the generated assertion, an \textit{SVA operator-based rechecking} is applied (Section~\ref{subsubsec:SVA-Op-based-Rechecking}). 
This refinement step verifies and corrects the usage of SVA operators, ensuring better alignment between the generated assertion and the intended property description. 
Finally, it outputs a refined assertion.

\subsubsection{Dynamic Splitting}
\label{subsubsec:Dynamic-Splitting}
addresses a key limitation of static splitting by distinguishing between code blocks and their related text. Instead of segmenting all content uniformly based on a fixed size threshold, our method constructs a specialized \textit{code database} designed to preserve the context surrounding each example.

The splitting process begins by parsing SystemVerilog textbooks to locate and isolate code snippets. 
Whenever a code block is detected, the system automatically gathers the paragraph immediately preceding the code and the paragraph immediately following it. 
These three elements, consisting of the paragraph before, the code snippet itself, and the paragraph after, are combined into a single chunk, called \textit{code-centric chunk}. 
Code-centric chunks are retained to construct the code database. 
Paragraphs not adjacent to a code snippet are not separately stored. 
By splitting content in this semantically-aware manner and focusing retrieval on code-centric contexts, dynamic splitting improves the relevance and precision of retrieved knowledge.
\subsubsection{HybridRetrieval}
\label{subsubsec:HybridRetrieval}
In addition to limitations in database construction, the retrieval process itself presents significant challenges when applying the basic RAG framework to SVA generation. Standard retrieval mechanisms based on global semantic similarity of the input prompt fail to capture fine-grained semantic embedded in natural language property specifications. Temporal relationships and operator-specific behaviors may be overlooked, resulting in the retrieval of incomplete or irrelevant contexts.

\begin{figure}
    \centering
    \includegraphics[width=\linewidth]{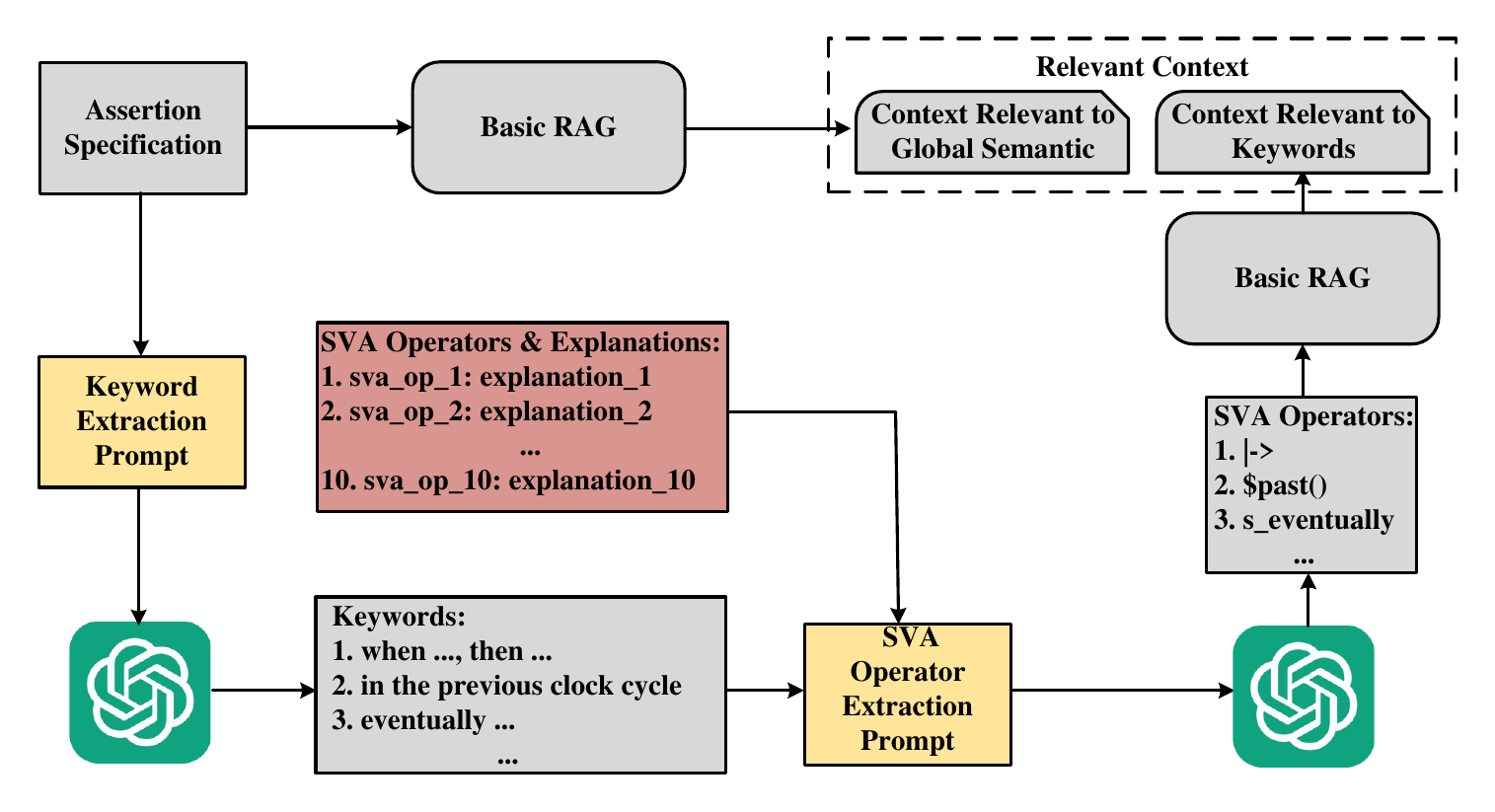}
    \caption{Flow of HybridRetrieval combines the global semantic retrieval and the keyword-guided operator retrieval.}
    \label{fig:HybridRetrieval}
\end{figure}

To address this issue, we propose \textit{HybridRetrieval}, a two-path retrieval mechanism that enhances context relevance by combining global semantic retrieval with keyword-guided operator retrieval. The overall flow is illustrated in Fig.~\ref{fig:HybridRetrieval}.
Given a natural language assertion specification, HybridRetrieval initiates two retrieval paths in parallel.
\begin{itemize}
    \item In the first path, a standard basic RAG retrieval is performed by embedding the entire input specification and retrieving chunks from the constructed code database based on global semantic similarity. 
    This step captures relevant contextual information corresponding to the global semantic of the property description.
    \item In the second path, the LLM is sequentially instructed with two custom prompts to extract fine-grain operator-related context. 
\end{itemize}

\begin{figure}
    \centering
    \includegraphics[width=\linewidth]{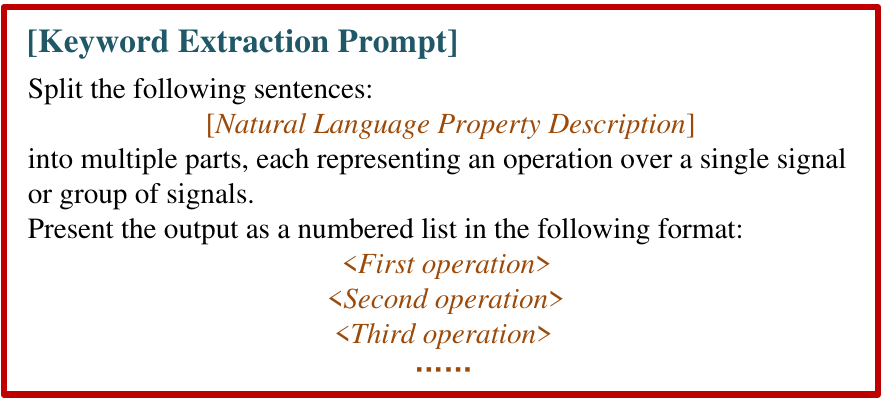}
    \caption{Keyword extraction prompt.}
    \label{fig:keyword_extraction_prompt}
\end{figure}

The first prompt is the \textit{Keyword Extraction Prompt}. 
It instructs the LLM to decompose the natural language description of the assertion into multiple parts, each representing an operation over a single signal or a group of signals.
Examples of such keywords include phrases like \texttt{in the previous clock cycle}, which indicate the presence of temporal relationships between signals.
The keyword extraction prompt is shown in Fig.~\ref{fig:keyword_extraction_prompt}.



\begin{figure}
    \centering
    \includegraphics[width=\linewidth]{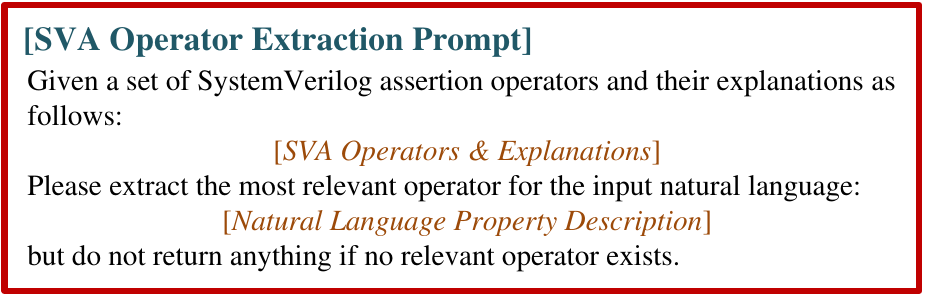}
    \caption{SVA operators extraction prompt.}
    \label{fig:sva_operators_extraction_prompt}
\end{figure}

The second prompt is the \textit{SVA Operator Extraction Prompt}, which is then used to instruct LLM to map each extracted keyword to the most relevant SVA operator.
To balance the complexity and accuracy of the mapping process, we prompt the LLM to select the most relevant operator from the $10$ SVA operators shown in Table~\ref{tab:sequence_sva_operators} and Table~\ref{tab:property_sva_operators}.
The SVA operator extraction prompt is shown in Fig.~\ref{fig:sva_operators_extraction_prompt}.



Once the relevant SVA operators are identified, a RAG retrieval is conducted. 
This retrieval  targets database chunks that mention or explain the identified operators, ensuring that the retrieved context is not only semantically relevant but also aligned with the SVA operator semantics necessary for correct assertion construction.

Finally, the outputs from the global semantic and the operator-guided retrievals are combined with the initial SVA generation prompt to enrich the prompt. 
By integrating both the property semantics and operator-specific contexts, HybridRetrieval can improve the quality and completeness of information for the LLM during SVA generation.

\subsubsection{SVA Operator-based Rechecking}
\label{subsubsec:SVA-Op-based-Rechecking}
Even with the improved retrieval through HybridRetrieval, LLMs may generate assertions that misuse SVA operators. 
Misinterpretations occur because retrieved materials, although relevant, may contain both helpful and noisy information. 
Certain SVA operators exhibit slight semantic differences, such as \texttt{|->} and \texttt{|=>}, that are difficult for LLMs to distinguish during initial generation, particularly regarding precise timing and logical behavior.

\begin{figure}
    \centering
    \includegraphics[width=\linewidth]{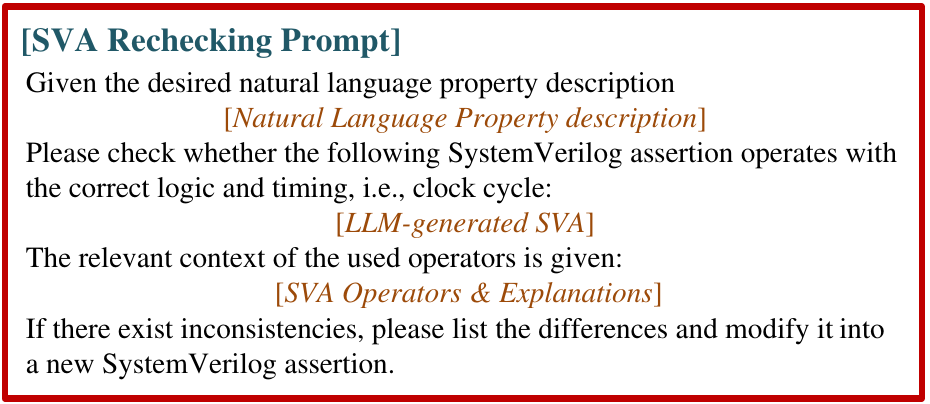}
    \caption{SVA checking prompt.}
    \label{fig:SVA_Checking_Prompt}
\end{figure}

To address these challenges, we propose an SVA operator-based rechecking flow. 
In this stage, the flow first extracts the list of operators present in the generated assertion. 
The relevant explanations for these operators are then retrieved from Table ~\ref{tab:sequence_sva_operators} and Table~\ref{tab:property_sva_operators}. 
Given these context, the \textit{SVA Rechecking Prompt} is applied to instruct the LLM to verify the correct use of SVA operators in the LLM-generated assertion. 
The SVA rechecking prompt is shown in Fig.~\ref{fig:SVA_Checking_Prompt}.
\subsection{Synthetic Finetuning Dataset for NL2SVA}
\label{subsec:synthetic-finetuning-dataset}
Although Section ~\ref{subsec:RAGSVAG} introduces a customized RAG framework for guiding LLMs to do NL2SVA, lightweight LLMs may struggle to follow the different chained prompts in that pipeline. 
Their shorter context windows can also truncate the combined prompt and retrieved context, leading to incomplete or incoherent output. 
To overcome these limits, we prepare a supervised fine-tuning dataset that teaches smaller models to perform the NL2SVA task directly, without relying on long, multi-stage prompts.

By scraping Verilog codes from $67$ hardware-design textbooks spanning digital logic, computer architecture, SoC implementation, and formal verification, and extracting every embedded SVA, we obtained $4070$ ground-truth assertions.
Each assertion is paired with a concise, one-sentence natural language explanation produced by prompting \textit{OpenAI o4-mini} with the raw SVA and natural language explanations of SVA operators (Table~\ref{tab:sequence_sva_operators} and Table~\ref{tab:property_sva_operators}).

The core feature of our fine-tuning dataset is a complete layer-by-layer derivation trace of each SVA from its corresponding natural language explanation. 
We refer to this trace as a \textit{Prompt-guided Explanation}. 
Section~\ref{subsec:SVA} describes the four layers of a concurrent SVA, and the prompt-guided explanation records how those layers are actually constructed for each individual case. 
The prompt-guided explanation is produced by OpenAI o4-mini. We feed each pair of the SVA and explanation to the LLM and instruct it to reconstruct the SVA from the explanation step by step as follows:
\begin{itemize}
    \item [(1)] Identify the top-level property SVA operator.
    \item [(2)] Split the natural language explanation into operand fragments that match the number of operands of the selected operator;
    \item [(3)] For each fragment, decide whether it represents a sequence or a property. If it is a sequence, translate it into a sequence expression directly; otherwise, recursively apply Steps $1$--$4$.
    \item [(4)] Combine the fragment expressions into the complete property expression using the selected property SVA operator.
    \item [(5)] Wrap the property expression in \texttt{assert property (...)}.
\end{itemize}
The detailed prompt for generating the prompt-guided explanation is shown in Appendix~\ref{subsec:prompt-guided-explanation-generation} and an example prompt-guided explanation is shown in Appendix~\ref{subsec:example-prompt-guided-explanation}.
\subsection{Evaluation Dataset}\label{subsec:Evaluation-Dataset}
\begin{figure}
    \centering
    \includegraphics[width=0.9\linewidth]{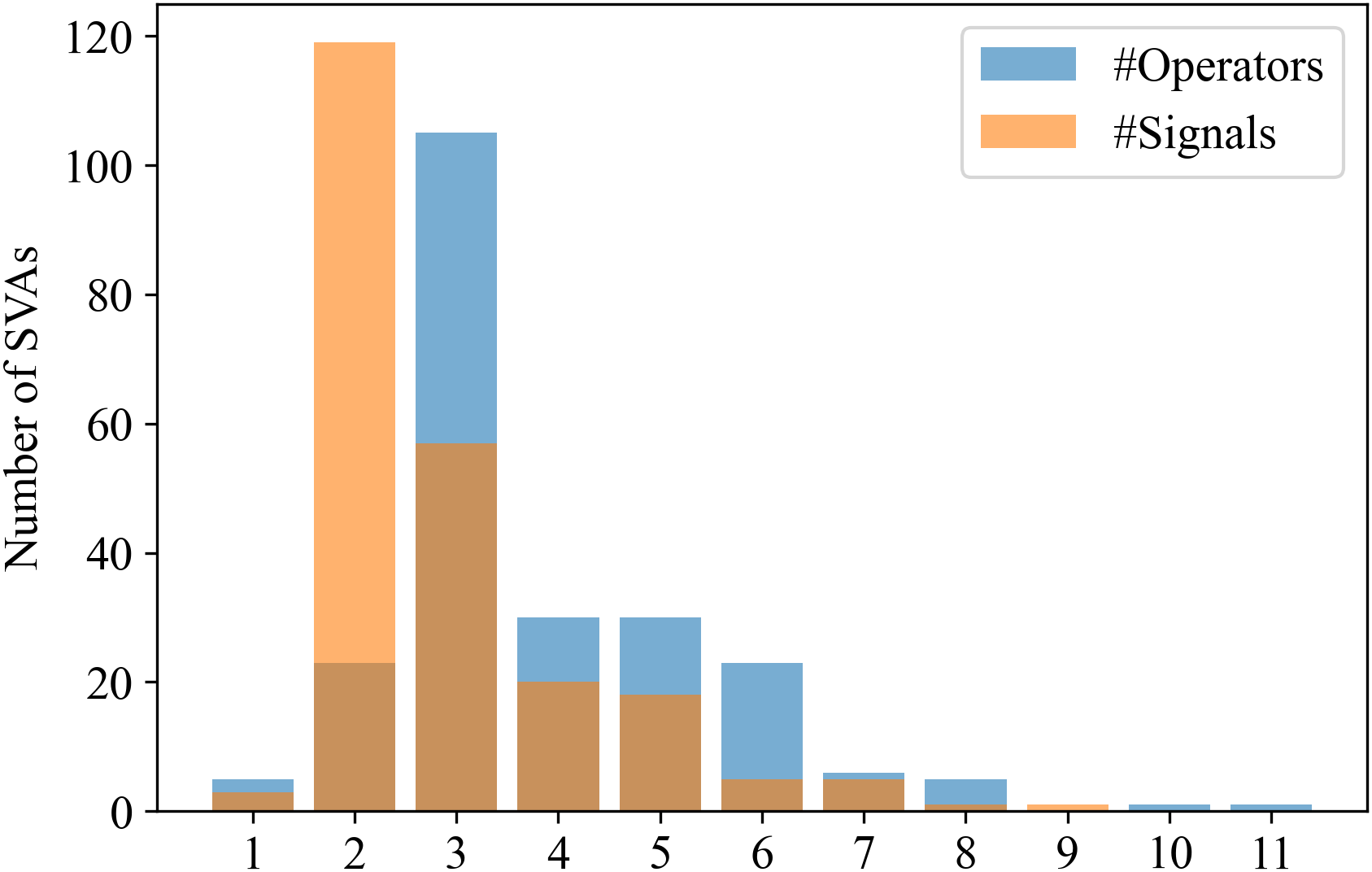}
    \caption{Number of SVA operators and signals in the $229$ SVAs.}
    \label{fig:sva-ops-signals}
\end{figure}

In constructing our evaluation dataset, we aim to ensure both realism and quality in the SVAs. 
To achieve this, we gather $40$ Verilog designs and $229$ concurrent SVAs from a combination of open-source cores and academic verification courses. 
In Fig.~\ref{fig:sva-ops-signals}, it summarizes the operator and signal counts for all $229$ SVAs, showing the complexity of each SVA and indicating that most of them are non-trivial.

Each of them is verified to ensure syntax and functionality correctness using FPV tool \textit{Cadence JasperGold}~\cite{cadencejaspergold}. 
The evaluation dataset provides the  natural language property for each SVA, which is generated manually.
Note that each explanation does not contain any detailed signals, which are translated as natural language descriptions.
In Appendix~\ref{subsec:Example-SVAs}, it shows $10$ example natural language properties from our evaluation dataset. 

%% file: Tex/Experiments.tex
\section{Experimental Results} \label{sec:exp}
\subsection{Experimental Setup}
\label{subsec:experimental-setup}
This experimental section first evaluates the effectiveness of our customized RAG framework for NL2SVA. 
To provide a comprehensive evaluation, we test  it across three LLMs: \textit{OpenAI GPT-4o-mini}, \textit{OpenAI CodeX} which has been fine-tuned on high-quality public code repositories, and \textit{DeepSeek-V3}.
We then evaluate the lightweight \textit{Qwen2.5-Coder-7B-Instruct} model fine-tuned on our synthetic prompt-guided dataset, comparing its performance with the base Qwen model.
We also apply our customized RAG techniques to Qwen and its fine-tuned variants and evaluate their performances.
For methods employing RAG, we retrieve relevant context from the database, which is constructed using $10$ hardware design and verification textbooks.

All experiments are conducted using our evaluation dataset. 
This dataset is the ground truth for evaluating the LLM-generated SVAs in terms of both syntax correctness and functionality match with the natural language property description. 
Two metrics are adopted during evaluation.
\begin{itemize}
    \item \textit{Syntax Correctness (SC)}: number of assertions that are syntactically correct.
    \item \textit{Functionality Match (FM)}: number of assertions that are functionally equivalent to the golden assertions in the evaluation dataset.
\end{itemize}
The two metrics are both derived by the FPV tool \textit{Cadence JasperGold} during our evaluation.
For SC, the FPV tool can directly check the syntax correctness of the LLM-generated SVAs.
For FM, we create the checking SVA by combining the property expression in the golden SVA with that in the LLM-generated SVA using SVA operator \texttt{iff}, as follows:
\begin{verbatim}
     assert property(gold_property_expression 
     iff LLM_property_expression);
\end{verbatim}
If the FPV tool cannot detect any counterexamples for the checking SVA, the two SVAs are functionally equivalent.

Given that our customized RAG framework has multiple components, we perform detailed evaluations on each of them. Specifically, Section~\ref{subsec:exp-evaluation-dynamic} evaluates the effectiveness of the dynamic splitting technique for database construction. 
Section~\ref{subsec:exp-evaluation-hybridretrieval} evaluates the proposed HybridRetrieval method for improving retrieval quality. 
Section~\ref{subsec:exp-evaluate-RAGSVAG} presents an overall evaluation of the customized RAG framework and a comparison with related works.
Finally, Section~\ref{subsec:evaluate-lightweight-llm} evaluates the lightweight Qwen2.5-Coder-7B-Instruct model and its fine-tuned variants using our synthetic dataset, both standalone and integrated with the techniques of our customized RAG framework.

\subsection{Evaluation of Dynamic Splitting Technique}
\label{subsec:exp-evaluation-dynamic}
In this subsection, we evaluate the effectiveness of the proposed dynamic splitting technique. 
We compare three methods: applying the base LLM without any retrieval augmentation (\textit{LLM}), applying retrieval augmentation with a database constructed via traditional static splitting (\textit{StaticRAG}), and applying retrieval augmentation with a database constructed using our dynamic splitting  (\textit{DynamicRAG}). 

\begin{figure}
    \centering
    \includegraphics[width=\linewidth]{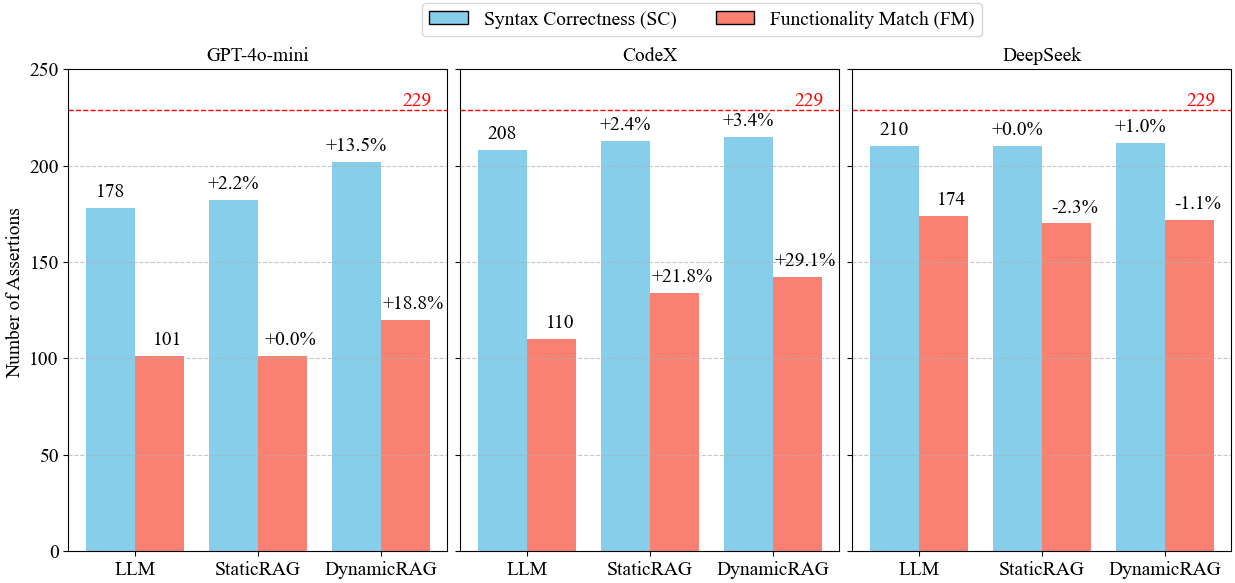}
    \caption{Evaluation of dynamic splitting technique.}
    \label{fig:eval-dynamic-splitting}
\end{figure}

The evaluation results are summarized in Fig.~\ref{fig:eval-dynamic-splitting}, where the improvement ratio over the basic LLM is shown. 
For GPT-4o-mini, DynamicRAG significantly improves FM by $18.8\%$ compared to LLM, while StaticRAG provides no improvement. 
In terms of SC, DynamicRAG result in $13.5\%$ improvement while StaticRAG only achieves a $2.2\%$ improvement.
When using CodeX, a similar trend is observed. DynamicRAG improves FM by $29.1\%$ over LLM, outperforming StaticRAG ($21.8\%$). 
However, both DynamicRAG and StaticRAG just achieve a slight improvement on SC over LLM.
For DeepSeek, DynamicRAG and StaticRAG even degrade the performance of FM and SC slightly compared to the basic LLM. 
This behavior may be due to DeepSeek’s strong built-in semantic understanding, making it less sensitive to retrieval augmentation quality.
Overall, these results demonstrate that dynamic splitting substantially can help improve the performance of SVA generation for the basic RAG framework.

\subsection{Evaluation of HybridRetrieval}
\label{subsec:exp-evaluation-hybridretrieval}
In this subsection, we evaluate the effectiveness of the proposed HybridRetrieval method. 
Recall that HybridRetrieval consists of the global semantic retrieval and keyword-guided retrieval.
Thus, we compare four methods: (1) using the basic LLM without retrieval augmentation (\textit{LLM}), (2) only global semantic retrieval (\textit{HR-P0}), (3) only keyword-guided retrieval (\textit{HR-P1}), and (4) the full HybridRetrieval (\textit{HR}). In all experiments, the dynamic splitting technique is applied to construct the retrieval database.

\begin{figure}
    \centering
    \includegraphics[width=\linewidth]{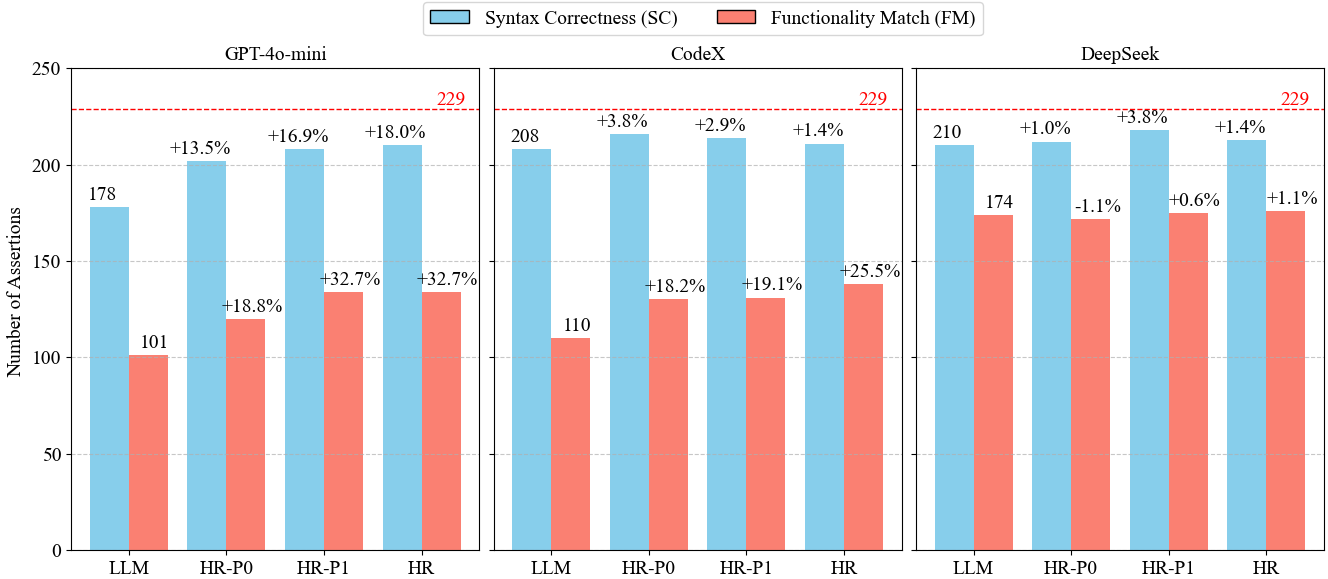}
    \caption{Evaluation of HybridRetrieval.}
    \label{fig:Evaluation-HybridRetrieval}
\end{figure}

The evaluation results are shown in Fig.~\ref{fig:Evaluation-HybridRetrieval}. 
When using GPT-4o-mini, applying HybridRetrieval (both HR-P0 and HR-P1 individually) improves FM substantially compared to the basic LLM. Specifically, HR-P0 improves FM by $18.8\%$ and HR-P1 improves it by $32.7\%$. 
Combining both techniques in HR maintains the $32.7\%$ improvement in FM.
For CodeX, similar trends are observed. HR-P0 and HR-P1 individually improve FM by $18.2\%$ and $19.1\%$, respectively, over the baseline, while the full HybridRetrieval (HR) achieves a $25.5\%$ improvement. 
When evaluating on DeepSeek, improvements are smaller due to the model's stronger built-in understanding ability.  HR still provides consistent FM improvements ($1.1\%$) compared to the baseline, while global semantic retrieval (HR-P0) shows slight $1.1\%$ FM reduction and keyword-guided retrieval (HR-P1) improves by $0.6\%$. 
The full HR yields the best FM performance.

\subsection{Evaluation and Comparison of Customized RAG Framework}
In this subsection, we present the overall evaluation of our full customized RAG framework (\textit{RAGSVAG}) using both HybridRetrieval and SVA operator-based rechecking. Specifically, we do comparison over: applying the plain LLM without any retrieval or augmentation (\textit{LLM}), applying our SVA operator-based rechecking flow without retrieval augmentation (\textit{SOR}), and applying the full RAGSVAG framework. 
For all methods involving RAG, our dynamic splitting technique and HybridRetrieval are employed.
Additionally, we compare our RAGSVAG with \textit{nl2spec}~\cite{Cosler23} and \textit{Spec2Assertion}~\cite{wu2025}.

\begin{figure}
    \centering
    \includegraphics[width=\linewidth]{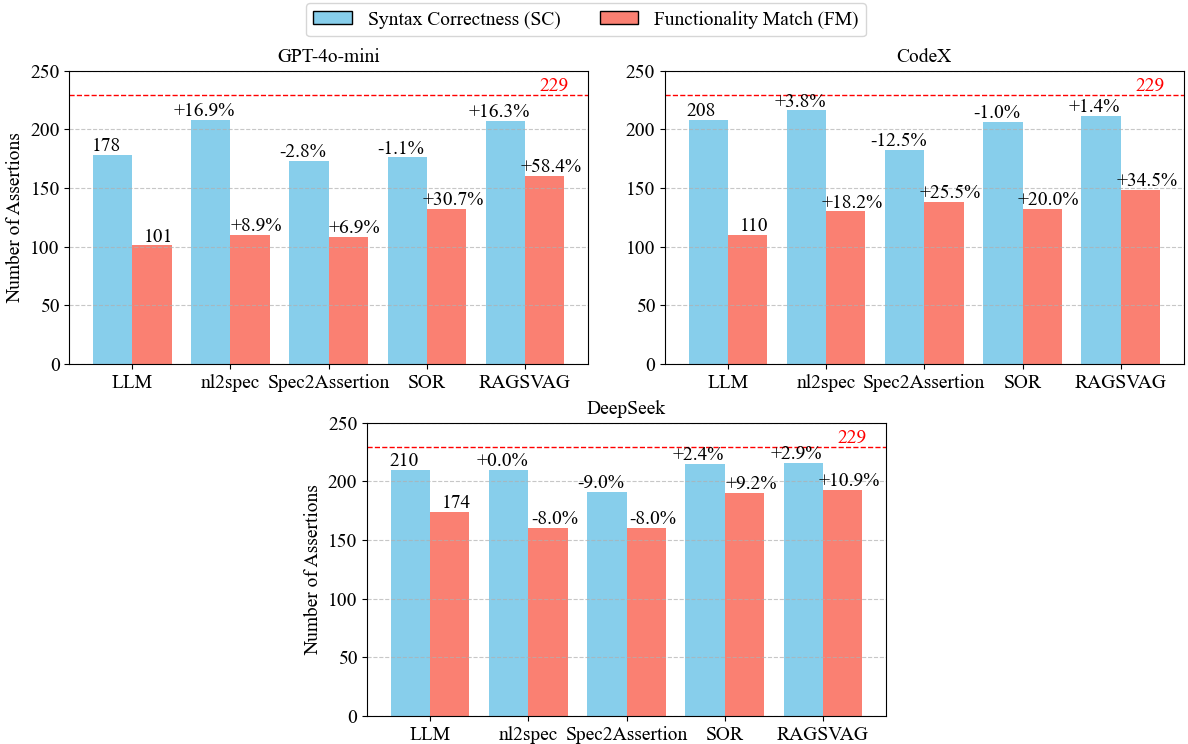}
    \caption{Evaluation and comparison of Customized RAG Framework.}
    \label{fig:evaluation-RAGSVAG}
\end{figure}

The results are shown in Fig.~\ref{fig:evaluation-RAGSVAG}. 
Across all the three tested LLMs, RAGSVAG consistently achieves the highest FM among all compared methods, while maintaining competitive SC.
For GPT-4o-mini, applying nl2spec improves FM by $8.9\%$ and applying Spec2Assertion improves FM by $6.9\%$, while SOR alone improves it by $30.7\%$. In contrast, RAGSVAG achieves a FM improvement of $58.4\%$, representing a substantial gain over both baseline and intermediate methods.
For CodeX, similar trends are observed. 
Applying nl2spec improves FM by $18.2\%$ and applying Spec2Assertion improves FM by $25.5\%$, and SOR improves it by $20.0\%$. RAGSVAG can achieve a $34.5\%$ improvement over FM,  outperforming the baselines.
For DeepSeek, which shows strong baseline performance, relative improvements are smaller.  RAGSVAG still achieves a FM improvement of $10.9\%$ compared to the basic LLM, while SOR provide smaller gains, and the nl2spec and Spec2Assertion actually underperform relative to the baseline.
In terms of SC, all methods show smaller improvement, indicating that retrieval and rechecking enhance the FC accuracy of the generated assertions rather than SC.
Overall, the results demonstrate the effectiveness of RAGSVAG in substantially improving the FM of LLM-generated SVAs.

\label{subsec:exp-evaluate-RAGSVAG}

\subsection{Evaluation of Finetuned Lightweight LLM}
\label{subsec:evaluate-lightweight-llm}
We evaluate the lightweight \textit{Qwen2.5-Coder-7B-Instruct} model~\cite{hui24} and two fine-tuned variants derived from our synthetic finetuning Dataset in Section~\ref{subsec:synthetic-finetuning-dataset}.  
The first variant, \textit{Qwen-Finetune}, is trained on plain (\emph{SVA, explanation}) pairs from our finetuning dataset following~\cite{Shahidzadeh24}.  
The second, \textit{Qwen-Prompt-Finetune}, is trained on (\emph{SVA, prompt-guided explanation}) pairs.  
The two fine-tuning tasks use supervised learning with \textit{Llama-Factory} \cite{zheng2024llamafactory}, a cosine learning-rate schedule, an initial rate of $8\times10^{-5}$, \texttt{bf16} precision and three epochs.  
Training the two variants on $8 \times$ A100 GPUs requires roughly two hours.

We integrate the techniques from our customised RAG framework into the base Qwen model and its two fine-tuned variants. 
We employ Qwen and its fine-tuned variants with a temperature of $0.6$, top\_p of $0.95$. Considering the format of our fine-tuning dataset and Qwen’s limited context window, we  feed the assertion specification together with the related signals and their natural language explanations, rather than the full Verilog code, into the model or RAG pipeline.

Table \ref{tab:qwen_results} reports results of SC and FM, the corresponding percentages relative to all $229$ SVAs, and the improvements achieved with respect to the base model.
The first column reports the results of the base Qwen. 
The model produces $161$ SC and $105$ FM SVAs, or $70.31 \%$ and $45.85 \%$ of the $229$ SVAs. 
Integrating HybridRetrieval  improves the two metrics to $198$ / $113$. 
Integrating our SVA operator-based rechecking achieves higher SC to $201$ but decreases FM to $101$, and integrating our customized RAG framework ends at $195$ / $101$. 
In short, the base model achieves similar SVA generation performance to that of GPT-4o-mini and benefits from HybridRetrieval, while the SVA operator-based rechecking components help improve SC.

The second column reports the results of Qwen-Finetune. This variant falls to $160$ / $87$, losing $0.62 \%$ in SC and $17.14 \%$ in FM compared with the base model, so we do not integrated any techniques into it. 
By contrast, the third column Qwen-Prompt-Finetune improves SC / FM to $206$ / $148$ ($89.96 \%$ / $64.63 \%$), a $27.95 \%$ SC gain and a $40.95 \%$ FM gain over the base model. 
When HybridRetrieval is integrated, the two metrics continue to be improved to $213$ / $167$, and they remain well above the baseline integrating SVA operator-based rechecking and the overall customized RAG framework. 
These results confirm that fine-tuning the lightweight Qwen model with prompt-guided explanations markedly improves SVA generation, and that integrating the HybridRetrieval component of our customized RAG framework provides an additional improvement.
\begin{table}[t]
\centering
\caption{Evaluation results of the base Qwen model, its finetuned variant, and its prompt-finetuned variant, each used together with our customised RAG framework.}
\label{tab:qwen_results}
\setlength{\tabcolsep}{1.5pt}
\renewcommand{\arraystretch}{1.35}
\begin{tabular}{@{}lcc cc cc@{}}
\toprule
\multirow{2}{*}{Methods} &
\multicolumn{2}{c}{Qwen} &
\multicolumn{2}{c}{Finetune-Qwen} &
\multicolumn{2}{c}{Prompt-Finetune-Qwen} \\ \cmidrule(lr){2-3}\cmidrule(lr){4-5}\cmidrule(l){6-7}
 & SC & FM & SC & FM & SC & FM \\ \midrule
LLM     & \tabincell{c}{161\\(70.31\%)} & \tabincell{c}{105\\(45.85\%)} & \tabincell{c}{160\\(69.87\%)} & \tabincell{c}{87\\(37.99\%)}  & \tabincell{c}{206\\(89.96\%)} & \tabincell{c}{148\\(64.63\%)} \\
\textit{Improv.} & --- & --- & -0.62\% & -17.14\% & +27.95\% & +40.95\% \\[1ex]
\hline
HR      & \tabincell{c}{198\\(86.46\%)} & \tabincell{c}{113\\(49.34\%)} & --- & --- & \tabincell{c}{\textbf{213}\\(\textbf{93.01\%})} & \tabincell{c}{\textbf{167}\\(\textbf{72.93\%})} \\
\textit{Improv.} & +22.98\% & +7.62\% & --- & --- & \textbf{+32.30}\% & \textbf{+59.05}\% \\[1ex]
\hline
SOR     & \tabincell{c}{201\\(87.77\%)} & \tabincell{c}{101\\(44.10\%)} & --- & --- & \tabincell{c}{208\\(90.83\%)} & \tabincell{c}{155\\(67.69\%)} \\
\textit{Improv.} & +24.84\% & -3.81\% & --- & --- & +29.19\% & +47.62\% \\[1ex]
\hline
RAGSVAG & \tabincell{c}{195\\(85.15\%)} & \tabincell{c}{101\\(44.10\%)} & --- & --- & \tabincell{c}{211\\(92.14\%)} & \tabincell{c}{164\\(71.62\%)} \\
\textit{Improv.} & +21.12\% & -3.81\% & --- & --- & +31.06\% & +56.19\% \\ \bottomrule
\end{tabular}
\end{table}

%% file: Tex/Conclusion.tex
\section{Conclusion} \label{sec:con}
In this paper, we present a customized RAG framework and a synthetic fine-tuning dataset of prompt-guided explanations to streamline SVA generation from natural language descriptions. The RAG framework integrates dynamic splitting, HybridRetrieval, and an operator-based rechecking flow for precise context retrieval and assertion validation, while the dataset provides layer-by-layer explanation of each SVA spanning the Boolean, sequence, property, and verification layers. 
We also construct the largest evaluation dataset for NL2SVA to date, with $40$ Verilog designs and $229$ formally verified assertions with detailed annotations.
Evaluation on GPT-4o-mini, CodeX, DeepSeek-V3, and a fine-tuned Qwen2.5-Coder-7B-Instruct model shows that combining our customized RAG framework with prompt-guided fine-tuning delivers marked improvements in syntax correctness and functionality match, enabling robust SVA synthesis with minimal manual effort.

%% file: Tex/Appendix.tex
\subsection{Example SVAs}
\label{subsec:Example-SVAs}
Across all $10$ examples in Table~\ref{tab:example_svas}, the SVAs generated by our customized RAG framework pass the functionality verification, meaning they accurately implement the property. In contrast, the SVAs generated by GPT-4o-mini exhibit functionality mismatches in every case.  Common issues include incorrect usage of temporal operators (such as overlapping implication operator \texttt{|->} and non-overlapping implications operator \texttt{|=>}), improper handling of past signal values, and failure to accurately capture the specified clock-cycle timing relationships.
\begin{table*}[ht]
    \centering
    \caption{Examples of SVAs generated by GPT-4o-mini and RAGSVAG.}
    \label{tab:example_svas}
    \begin{tabular}{|c|c|p{3.8cm}|p{5.4cm}|p{5.4cm}|}
        \hline
        \textbf{Design} & \textbf{Lines} & \textbf{Natural Language Explanation} & \textbf{GPT-4o-mini SVA} & \textbf{RAGSVAG SVA} \\
        \hline
        ff & $10$ & When enable signal is set (1), output equals input from previous cycle in next cycle. & \texttt{assert property (@(posedge clk) disable iff (rst) (en == 1'b1 |=> out == in[1]));} & \texttt{assert property (@(posedge clk) disable iff (rst) (en |=> (out == \$past(in))));} \\
        \hline
        PWM & $18$ & Eventually, the inversion of the output becomes true. & \texttt{assert property (@(posedge clk) (pulse |-> eventually (!pulse)));} & \texttt{assert property (@(posedge clk) (s\_eventually (!pulse)));}\\
        \hline
        register & $20$ & When enable reset (0), output equals last output in next cycle. & \texttt{assert property (@(posedge clk) disable iff (rst) (en == 0 |-> out == out[1]));} & \texttt{assert property (@(posedge clk) disable iff (rst) (en == 0 |=> (out == \$past(out))));} \\
        \hline
        simple\_req\_ack & $31$ & When request signal is detected to have a rising edge, then after a delay of $2$ clock cycles the acknowledge signal is detected to have a rising edge. & \texttt{assert property (@(posedge clk) disable iff(!rst\_n) (req\_ff in \{1'b0, 1'b1\} during [2] |=> (ack\_ff == 1'b1)));} & \texttt{assert property (@(posedge clk) disable iff(!rst\_n) (\$rose(req) |=> \#\#2 \$rose(ack)));} \\
        \hline
        or1200\_if & $194$ & When pipeline flush signal is reset ($0$) and save instruction condition is true, then saved flag equals 1 in the next clock cycle. & \texttt{assert property (@(posedge clk) disable iff (rst) (if\_flushpipe == 0 \&\& save\_insn |-> saved));} & \texttt{assert property (@(posedge clk) disable iff (rst) (if\_flushpipe == 0 \&\& save\_insn |=> (saved == 1)));} \\
        \hline
        arbiter & $231$ & When the first bit of grant output is asserted and in the previous clock cycle the arbitration type selector signal was equal to $0$, then the first bit of request input signal from the previous clock cycle must be asserted in the current clock cycle. & \texttt{assert property (@(posedge clk) disable iff (\textasciitilde rst\_n) (gnt[0] \&\& arb\_type == 0 |-> req[0]));} & \texttt{assert property (@(posedge clk) disable iff (\textasciitilde rst\_n) ((gnt[0] \&\& \$past(arb\_type) == 0) |-> \$past(req[0])));} \\
        \hline
        a25\_wishbone & $447$ & When the wishbone bus state equals \texttt{WB\_BURST1} and the acknowledgement signal is asserted, then the wishbone bus state equals \texttt{WB\_BURST2} in the next clock cycle. & \texttt{assert property (@(posedge i\_clk) (wishbone\_st == WB\_BURST1 \&\& i\_wb\_ack |-> (wishbone\_st == WB\_BURST2)));} & \texttt{assert property (@(posedge i\_clk) ((wishbone\_st == WB\_BURST1 \&\& i\_wb\_ack) |=> (wishbone\_st == WB\_BURST2)));} \\
        \hline
        control\_unit & $602$ & When the signal that indicates the beginning of an AES operation equals 0, then the signal used to initialize the key generation process equals 0 in the current clock cycle. & \texttt{assert property (@(posedge clk) (start == 0 |=> key\_init == 0));} & \texttt{assert property (@(posedge clk) (start == 1'b0 |-> key\_init == 1'b0));} \\
        \hline
        or1200\_ctrl & $1225$ & When the freeze signal for the EX stage is reset ($0$) and either the freeze signal for the ID stage is set ($1$) or the pipeline flush signal for the EX stage is set ($1$), then the signal indicating a MAC read operation in the EX stage equals 0 in the next clock cycle. & \texttt{assert property (@(posedge clk) disable iff (rst) (!ex\_freeze \&\& (id\_freeze || ex\_flushpipe) |-> (ex\_macrc\_op == 0)));} & \texttt{assert property (@(posedge clk) disable iff (rst) ((!ex\_freeze \&\& (id\_freeze || ex\_flushpipe)) |=> (ex\_mac\_op == 0)));} \\
        \hline
        module\_i2c & $3382$ & When the FIFO receiver empty indicator equals $1$, then the FIFO receiver empty output equals $1$ in the current clock cycle. & \texttt{assert property (@(posedge PCLK) (fifo\_rx\_f\_empty == 1'b1 |-> fifo\_rx\_wr\_en == 1'b1));} & \texttt{assert property (@(posedge PCLK) (fifo\_rx\_f\_empty == 1'b1 |-> RX\_EMPTY == 1'b1));} \\
        \hline
    \end{tabular}
\end{table*}

\subsection{Prompt-guided Explanation Generation}
\label{subsec:prompt-guided-explanation-generation}
\begin{figure}
    \centering
    \includegraphics[width=0.6\linewidth]{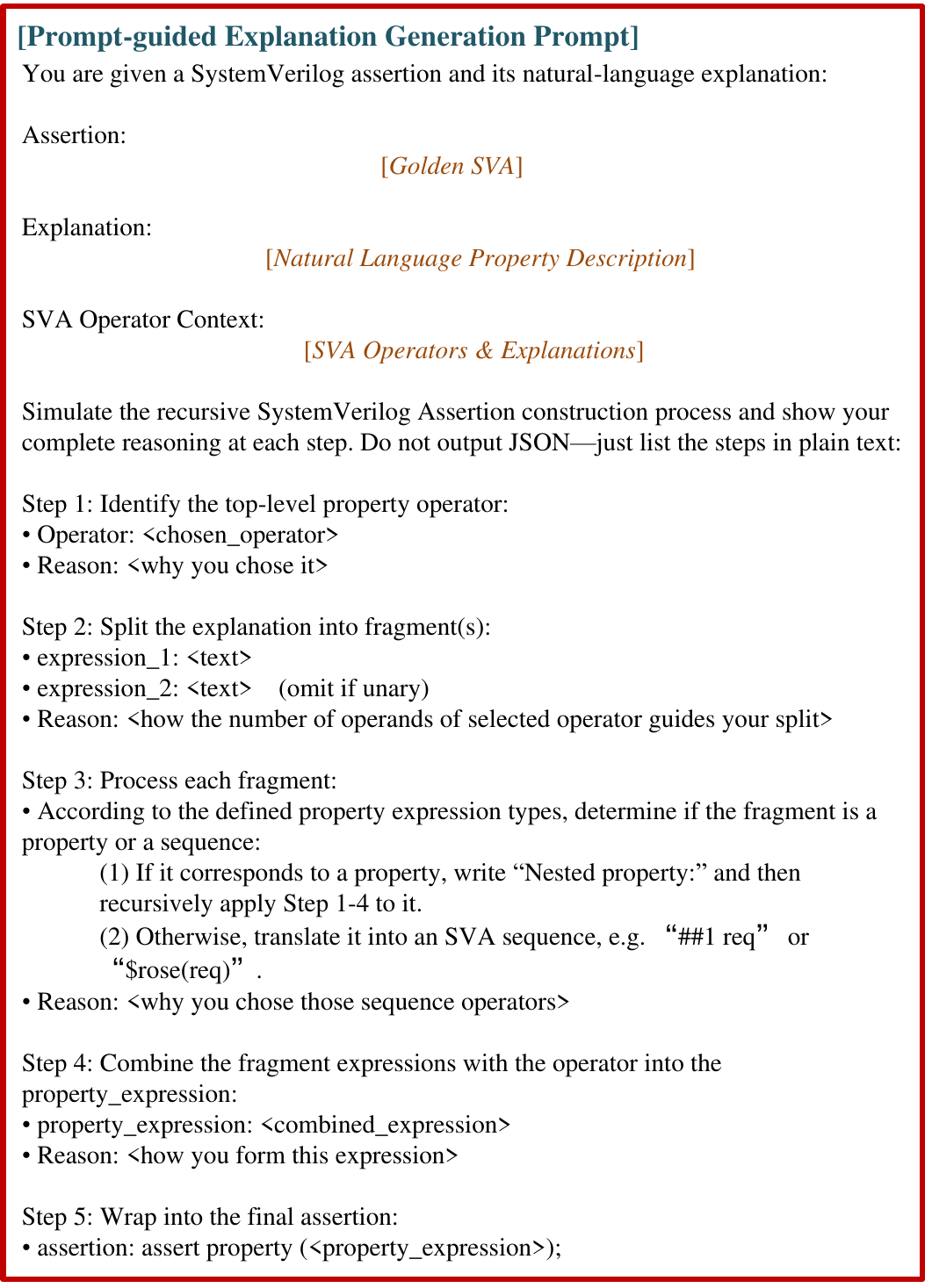}
    \caption{Prompt for generating the prompt-guided explanation.}
    \label{fig:prompt-guided-explanation-generation}
\end{figure}

The prompt for generating the prompt-guided explanation of the given golden SVA and natural language explanation is shown in Fig.~\ref{fig:prompt-guided-explanation-generation}. 

\subsection{Example Prompt-guided Explanation}
\label{subsec:example-prompt-guided-explanation}
\begin{figure}
    \centering
    \includegraphics[width=0.6\linewidth]{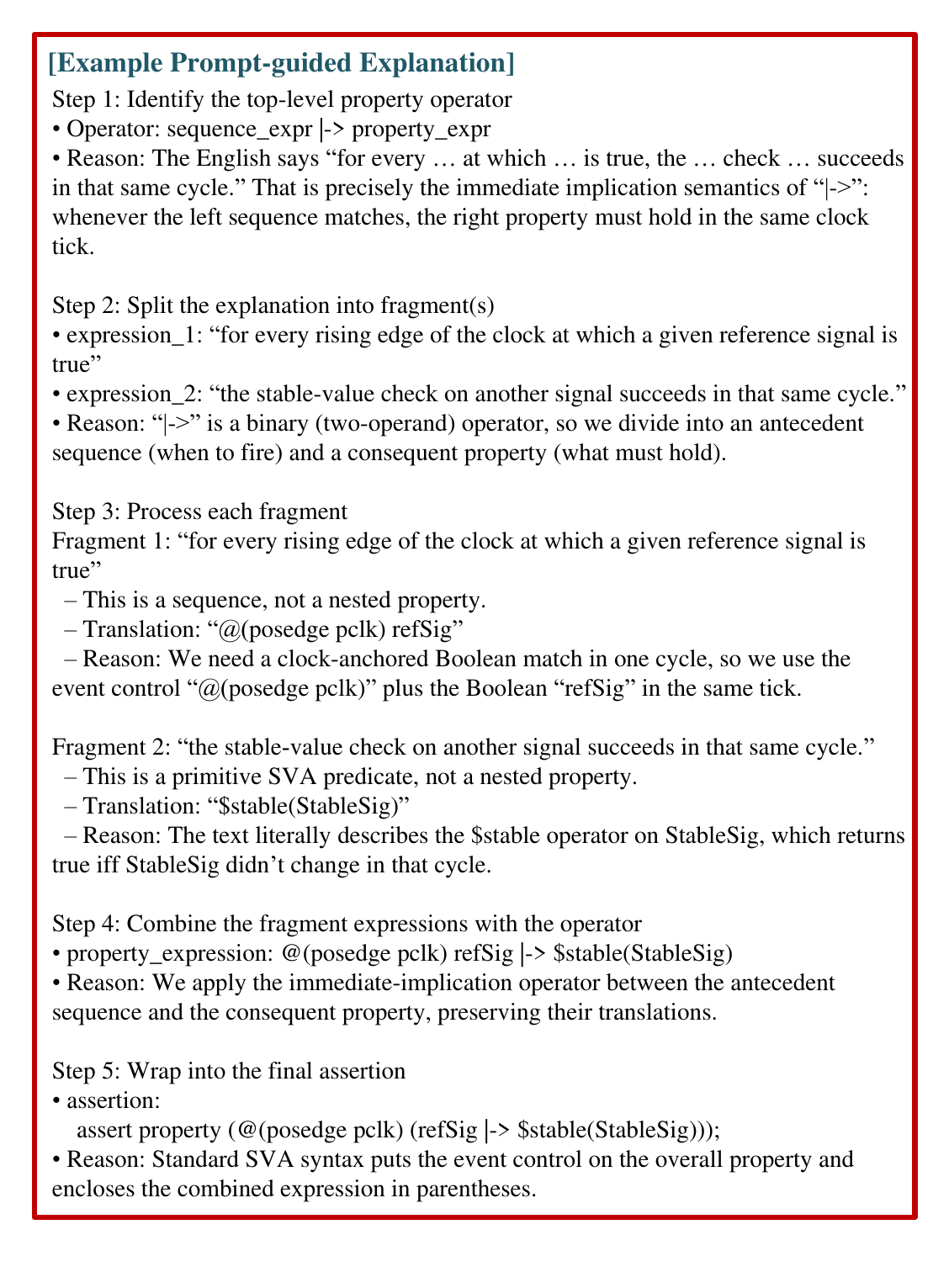}
    \caption{An example prompt-guided explanation.}
    \label{fig:example-prompt-guided-explanation}
\end{figure}

Golden SVA:
\begin{verbatim}
    @(posedge pclk) refSig |-> $stable(StableSig)
\end{verbatim}
Natural language explanation:

A property expression that for every rising edge of the clock at which a given reference signal \texttt{refSig} is true, the stable‐value check on another signal \texttt{StableSig} succeeds in that same cycle.
\\
\\
The corresponding prompt-guided explanation is shown in Fig.~\ref{fig:example-prompt-guided-explanation}, which is generated by OpenAI o4-mini.

%% file: main.bbl
\begin{thebibliography}{10}
\providecommand{\url}[1]{#1}
\csname url@samestyle\endcsname
\providecommand{\newblock}{\relax}
\providecommand{\bibinfo}[2]{#2}
\providecommand{\BIBentrySTDinterwordspacing}{\spaceskip=0pt\relax}
\providecommand{\BIBentryALTinterwordstretchfactor}{4}
\providecommand{\BIBentryALTinterwordspacing}{\spaceskip=\fontdimen2\font plus
\BIBentryALTinterwordstretchfactor\fontdimen3\font minus \fontdimen4\font\relax}
\providecommand{\BIBforeignlanguage}[2]{{%
\expandafter\ifx\csname l@#1\endcsname\relax
\typeout{** WARNING: IEEEtran.bst: No hyphenation pattern has been}%
\typeout{** loaded for the language `#1'. Using the pattern for}%
\typeout{** the default language instead.}%
\else
\language=\csname l@#1\endcsname
\fi
#2}}
\providecommand{\BIBdecl}{\relax}
\BIBdecl

\bibitem{Witharana22}
H.~Witharana \emph{et~al.}, ``A survey on assertion-based hardware verification,'' \emph{ACM Comput. Surv.}, vol.~54, no. 11s, pp. 1--33, 2022.

\bibitem{SystemVerilog18}
``Ieee standard for systemverilog--unified hardware design, specification, and verification language,'' \emph{IEEE Std 1800--2017}, pp. 1--1315, 2018.

\bibitem{Vasudevan10}
S.~Vasudevan, D.~Sheridan, S.~Patel, D.~Tcheng, B.~Tuohy, and D.~Johnson, ``{GoldMine}: automatic assertion generation using data mining and static analysis,'' in \emph{DATE}, 2010, pp. 626--629.

\bibitem{Witharana23}
H.~Witharana, A.~Jayasena, A.~Whigham, and P.~Mishra, ``Automated generation of security assertions for rtl models,'' \emph{J. Emerg. Technol. Comput. Syst.}, vol.~19, no.~1, 2023.

\bibitem{Radu24}
V.~Radu \emph{et~al.}, ``Generative {AI} assertions in {UVM}-based system verilog functional verification,'' \emph{Systems}, vol.~12, no.~10, 2024.

\bibitem{Mali24}
B.~Mali, K.~Maddala, V.~Gupta, S.~Reddy, C.~Karfa, and R.~Karri, ``{ChIRAAG}: {ChatGPT} informed rapid and automated assertion generation,'' in \emph{IEEE Computer Society Annual Symposium on VLSI}, 2024, pp. 680--683.

\bibitem{Yan25}
Z.~Yan \emph{et~al.}, ``{AssertLLM}: Generating hardware verification assertions from design specifications via multi-llms,'' in \emph{ASP-DAC}, 2025, pp. 614--621.

\bibitem{parthasarathy2021spectosva}
G.~Parthasarathy, S.~Nanda, P.~Choudhary, and P.~Patil, ``{SpecToSVA}: Circuit specification document to systemverilog assertion translation,'' in \emph{Second Document Intelligence Workshop at KDD}, 2021.

\bibitem{Meng24}
X.~Meng \emph{et~al.}, ``{NSPG}: Natural language processing-based security property generator for hardware security assurance,'' in \emph{DAC}, 2024, pp. 1--6.

\bibitem{Harris16}
C.~B. Harris and I.~G. Harris, ``{GLAsT}: Learning formal grammars to translate natural language specifications into hardware assertions,'' in \emph{DATE}, 2016, pp. 966--971.

\bibitem{Krishnamurthy19}
R.~Krishnamurthy and M.~S. Hsiao, ``{EASE}: Enabling hardware assertion synthesis from english,'' in \emph{RuleML+RR}, 2019, pp. 82--96.

\bibitem{Cosler23}
M.~Cosler, C.~Hahn, D.~Mendoza, F.~Schmitt, and C.~Trippel, ``nl2spec: Interactively translating unstructured natural language to temporal logics withlarge language models,'' in \emph{Computer Aided Verification}, 2023, pp. 383--396.

\bibitem{wu2025}
F.~Wu \emph{et~al.}, ``{Spec2Assertion}: Automatic pre-{RTL} assertion generation using large language models with progressive regularization,'' \emph{arXiv preprint arXiv:2505.07995}, 2025.

\bibitem{Shahidzadeh24}
M.~Shahidzadeh, B.~Ghavami, S.~Wilton, and L.~Shannon, ``Automatic high-quality verilog assertion generation through subtask-focused fine-tuned {LLMs} and iterative prompting,'' \emph{arXiv}, 2024.

\bibitem{IEEEStd}
``Ieee standard for systemverilog--unified hardware design, specification, and verification language,'' \emph{IEEE Std 1800-2023 (Revision of IEEE Std 1800-2017)}, pp. 1--1354, 2024.

\bibitem{Lewis20}
P.~Lewis \emph{et~al.}, ``{R}etrieval-augmented generation for knowledge-intensive {NLP} tasks,'' in \emph{Advances in neural information processing systems}, 2020, pp. 9459--9474.

\bibitem{Koziolek24}
H.~Koziolek \emph{et~al.}, ``{LLM}-based and retrieval-augmented control code generation,'' in \emph{International Workshop on Large Language Models for Code}, 2024, pp. 22--29.

\bibitem{xiong2024}
G.~Xiong, Q.~Jin, Z.~Lu, and A.~Zhang, ``Benchmarking retrieval-augmented generation for medicine,'' in \emph{Association for Computational Linguistics}, 2024, pp. 6233--6251.

\bibitem{Setty24}
S.~Setty, H.~Thakkar, A.~Lee, E.~Chung, and N.~Vidra, ``Improving retrieval for {RAG} based question answering models on financial documents,'' \emph{arXiv}, 2024.

\bibitem{alonso2024}
N.~Alonso, T.~Figliolia, A.~Ndirango, and B.~Millidge, ``Toward conversational agents with context and time sensitive long-term memory,'' \emph{arXiv}, 2024.

\bibitem{singh24}
I.~Singh \emph{et~al.}, ``{ChunkRAG}: Novel {LLM}-chunk filtering method for rag systems,'' \emph{arXiv}, 2024.

\bibitem{chang24n}
C.~Chang \emph{et~al.}, ``{MAIN-RAG}: {Multi-Agent Filtering Retrieval-Augmented Generation},'' \emph{arXiv}, 2024.

\bibitem{dong24}
J.~Dong, B.~Fatemi, B.~Perozzi, L.~Yang, and A.~Tsitsulin, ``{Don't forget to connect! improving rag with graph-based reranking},'' \emph{arXiv}, 2024.

\bibitem{Vaishnavi25}
\BIBentryALTinterwordspacing
P.~Vaishnavi, N.~Deeksha, D.~Soham, and P.~Debjit, ``{AssertionBench}: A benchmark to evaluate large-language models for assertion generation,'' 2025. [Online]. Available: \url{https://arxiv.org/abs/2406.18627}
\BIBentrySTDinterwordspacing

\bibitem{Minwoo24}
\BIBentryALTinterwordspacing
K.~Minwoo \emph{et~al.}, ``{FVEval}: Understanding language model capabilities in formal verification of digital hardware,'' 2024. [Online]. Available: \url{https://arxiv.org/abs/2410.23299}
\BIBentrySTDinterwordspacing

\bibitem{cadencejaspergold}
\BIBentryALTinterwordspacing
C.~D. Systems, ``{Cadence JasperGold Formal Verification Platform},'' 2023. [Online]. Available: \url{https://www.cadence.com/}
\BIBentrySTDinterwordspacing

\bibitem{hui24}
B.~Hui \emph{et~al.}, ``{Qwen2. 5-coder} technical report,'' \emph{arXiv}, 2024.

\bibitem{zheng2024llamafactory}
Y.~Zheng \emph{et~al.}, ``{Llamafactory}: Unified efficient fine-tuning of 100+ language models,'' in \emph{ACL}, 2024.

\end{thebibliography}
